\documentclass{article}

\usepackage{microtype}
\usepackage{graphicx}
\usepackage{subfigure}
\usepackage{booktabs} 

\usepackage{hyperref}

\usepackage[accepted]{icml2025}

\usepackage{amsmath}
\usepackage{amssymb}
\usepackage{mathtools}
\usepackage{amsthm}

\usepackage[capitalize,noabbrev]{cleveref}

\theoremstyle{plain}

\theoremstyle{definition}

\theoremstyle{remark}

\usepackage[textsize=tiny]{todonotes}

\usepackage{multirow}
\usepackage{booktabs}
\usepackage{colortbl}
\usepackage{pifont}
\usepackage{bm}
\usepackage{enumitem}
\icmltitlerunning{Knowledge Rectification for Camouflaged Object Detection: Unlocking Insights from Low-Quality Data}

\begin{document}

\twocolumn[
\icmltitle{Knowledge Rectification for Camouflaged Object Detection: Unlocking Insights from Low-Quality Data}



\icmlsetsymbol{equal}{*}

\begin{icmlauthorlist}
\icmlauthor{Juwei Guan}{yyy}
\icmlauthor{Xiaolin Fang}{yyy}
\icmlauthor{Donghyun Kim}{comp}
\icmlauthor{Haotian Gong}{yyy}
\icmlauthor{Tongxin Zhu}{yyy}
\icmlauthor{Zhen Ling}{yyy}
\icmlauthor{Ming Yang}{yyy}
\end{icmlauthorlist}

\icmlaffiliation{yyy}{Southeast, University}
\icmlaffiliation{comp}{Korea University}

\icmlcorrespondingauthor{Xiaolin Fang}{xiaolin@seu.edu}

\icmlkeywords{Machine Learning, ICML}

\vskip 0.3in
]



\printAffiliationsAndNotice{\icmlEqualContribution} 

\begin{abstract}
Low-quality data often suffer from insufficient image details, introducing an extra implicit aspect of camouflage that complicates camouflaged object detection (COD). Existing COD methods focus primarily on high-quality data, overlooking the challenges posed by low-quality data, which leads to significant performance degradation. Therefore, we propose KRNet, the first framework explicitly designed for COD on low-quality data. KRNet presents a Leader-Follower framework where the Leader extracts dual gold-standard distributions: conditional and hybrid, from high-quality data to drive the Follower in rectifying knowledge learned from low-quality data. The framework further benefits from a cross-consistency strategy that improves the rectification of these distributions and a time-dependent conditional encoder that enriches the distribution diversity. Extensive experiments on benchmark datasets demonstrate that KRNet outperforms state-of-the-art COD methods and super-resolution-assisted COD approaches, proving its effectiveness in tackling the challenges of low-quality data in COD.
\end{abstract}

\section{Introduction}
Many animals in the wild employ strategies to adjust their appearance or select environments that allow them to blend seamlessly with their surroundings. Camouflaged object detection (COD) is a task aimed at breaking this camouflage to identify objects hidden within their environments. This technique finds broad applications in areas such as wildlife conservation and performance art \cite{le2019anabranch, fan2020camouflaged}. 
\begin{figure}[t!]
	\centering
	\includegraphics[width=0.95\linewidth]{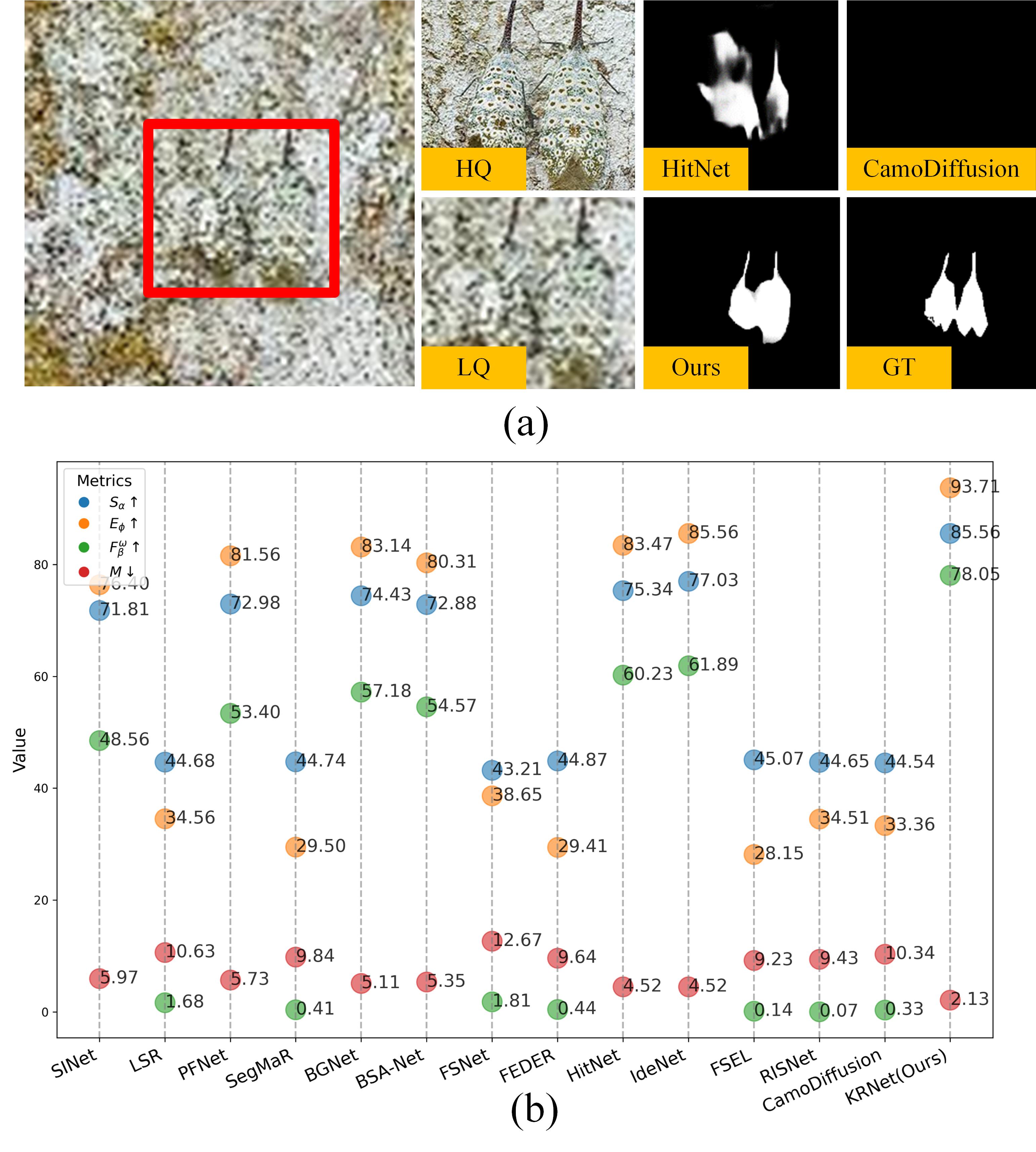}
        \vspace{-10pt}
	\caption{(a) Comparison of different COD methods on low-quality data. HitNet and CamoDiffusion fail to detect camouflaged object from low-quality image (LQ). HQ refers to high-quality image, and GT denotes the manually annotated masks.
(b) A performance comparison of our KRNet with other SOTA methods on COD10K.}
	\label{Fig:intro}
    \vspace{-1.2mm}
\end{figure}
Building upon previous contributions in collecting and annotating high-quality image datasets \cite{le2019anabranch, fan2020camouflaged, lv2021simultaneously}, learning-based COD methods \cite{sun2022boundary, guan2024idenet} have achieved considerable progress. These methods, in general, leverage the advancement of feature extractors and increasingly sophisticated camouflage-breaking strategies to effectively extract beneficial camouflage cues from the data, driving continuous progress of recognition performance. 

However, under low-quality data conditions, as illustrated in Figure \ref{Fig:intro}(a), the lack of critical cues such as edges and textures often leads to the collapse of these methods to detect camouflaged objects or a notable decline in performance. As shown in Table \ref{Tab:degra_per}, the state-of-the-art (SOTA) methods HitNet \cite{hu2023high} and CamoDiffusion \cite{chen2024camodiffusion}, whether trained on high-quality or low-quality data, suffer from significant performance degradation when evaluated on low-quality test datasets. Therefore, unlocking the detection of camouflaged objects in low-quality data, an area previously overlooked by current methods, and prompting the adaptability of COD models to low-quality settings, remains a critical challenge for advancing the field.

To the best of our knowledge, we are the first to present the novel concept of COD with low-quality data to overcome the performance limitations faced by traditional COD methods when attempting to break camouflage in conditions with insufficient information. Drawing inspiration from knowledge distillation \cite{hinton2015distilling}, we propose KRNet, the first method designed for COD in low-quality data. KRNet employs a knowledge rectification strategy that corrects the knowledge learned from low-quality data, allowing the model to uncover more contributory and informative cues for effective camouflage detection. Our KRNet demonstrates superior performance, as shown in Figure \ref{Fig:intro}(b), significantly outperforming existing COD methods and effectively narrowing the performance gap between low-quality and high-quality data.

\begin{figure*}[tp!]
	\centering
	\includegraphics[width=0.8\linewidth]{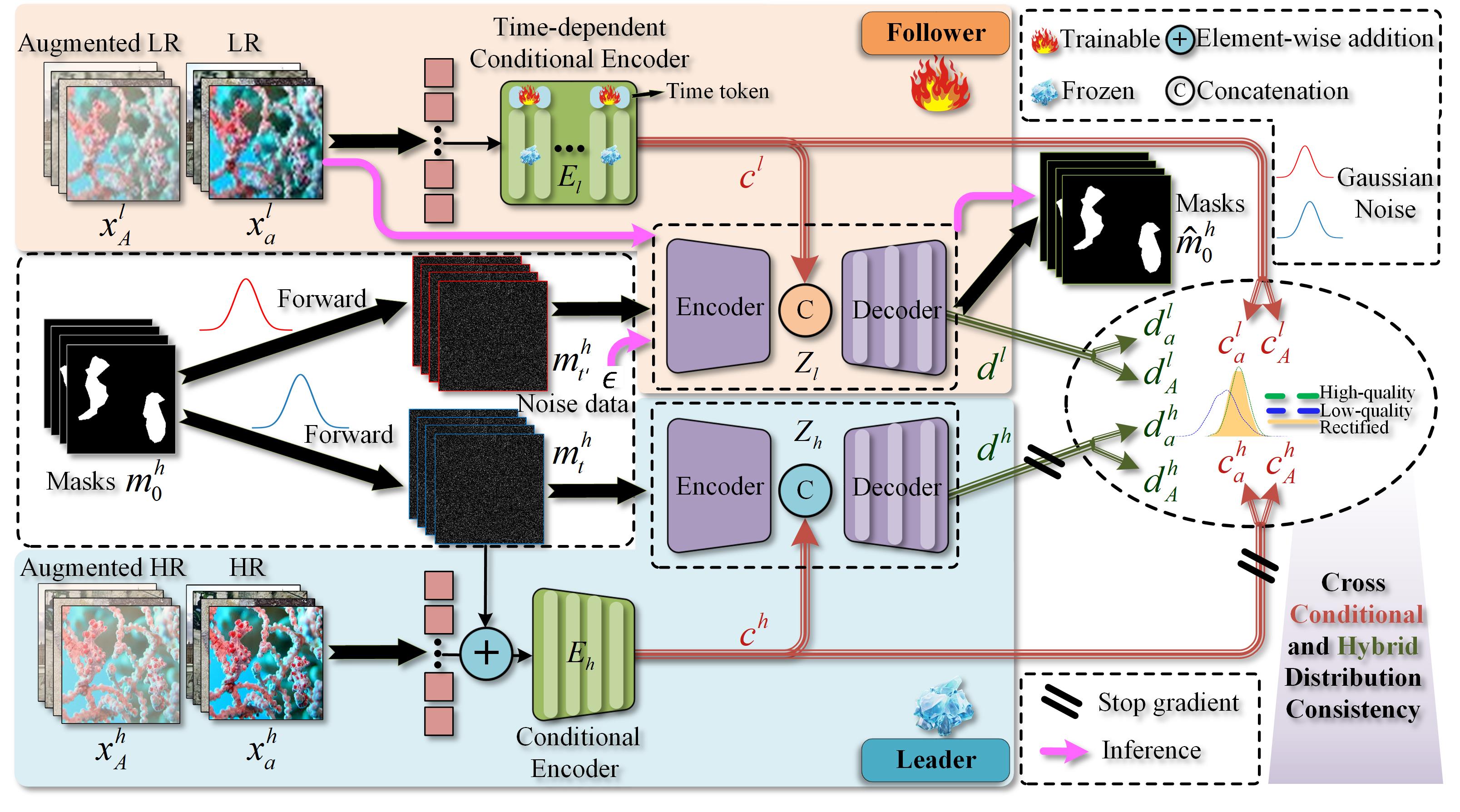}
    \vspace{-2pt}
	\caption{The framework of the proposed KRNet. KRNet consists of the Leader-Follower model based on a conditional diffusion model. The frozen Leader extracts stable conditional ($\bm{c}$) and hybrid ($\bm{d}$) distributions from high-quality data, while the Follower operates these distributions as gold standards to rectify knowledge learned from low-quality data. For inference, high-quality segmentation is achieved using only low-quality data and randomly sampled Gaussian noise.}
	\label{Fig:framework}
    \vspace{-1.5mm}
\end{figure*}

\begin{table}[t]
\centering
\vspace{-2mm}
\caption{Performance comparison under different quality data. 'H' is high-quality data, 'L' is low-quality data, 'H$\rightarrow$L' is train on high-quality data and test on low-quality data.}
\renewcommand{\arraystretch}{0.7}
\setlength{\tabcolsep}{3.4pt}{
    \begin{tabular}{ccc|c|c|c}
    \toprule
    \multirow{2}{*}{\textbf{Methods}}&\multirow{1}{*}{\textbf{Train\&}} & \multicolumn{4}{c}{\textbf{COD10K}}  \\
    \cmidrule(lr){3-6}
    & \textbf{Test}& $S_\alpha$$\uparrow$& $E_\phi$$\uparrow$& $F_{\beta}^{\omega}$$\uparrow$& $M$$\downarrow$ \\
    \midrule
    \multirow{3}{*}{\textbf{HitNet}}&H$\rightarrow$H&82.55&90.01&72.63&3.02\\
    &H$\rightarrow$L&73.03&	80.86&	56.07&5.96\\
    &L$\rightarrow$L&75.34&83.47&60.23&4.52\\
    \midrule
    \multirow{3}{*}{\textbf{CamoDiffusion}}&H$\rightarrow$H&87.90&94.42&81.20&1.99\\
    &H$\rightarrow$L&77.26&	84.66&	63.17&	6.72\\
    &L$\rightarrow$L&44.54&	33.36&	0.33&10.34\\
    \bottomrule
    \end{tabular}}
\label{Tab:degra_per}
\vspace{-1.8mm}
\end{table}
Our KRNet is based on a Leader-Follower framework, where the Leader learns camouflage-breaking knowledge from high-quality data, while the Follower learns knowledge constrained by limited information from low-quality data. Thus, two key issues arise: (1) how to design the architectures for both the Leader and Follower and (2) how to perform knowledge distillation between the Leader and Follower to correct the knowledge learned by the Follower. Most COD methods, with their diverse camouflage-breaking strategies and complex information interplay, require extensive exploration and a meticulous design for knowledge rectification, making them challenging to serve as benchmarks for knowledge learners. However, we find that the conditional diffusion model \cite{saharia2022image, chen2024camodiffusion}, when conditioned on images, can significantly simplify the process of knowledge rectification. Specifically, by structuring both the Leader and Follower around the conditional diffusion model, knowledge rectification is performed by leveraging the conditional signal learned by the Leader as the gold standard to refine that learned by the Follower, as illustrated in Figure \ref{Fig:framework}. This strategy is referred to as knowledge rectification through conditional distribution consistency (CDC). Furthermore, the conditional signal is involved in the denoising process, prompting us to propose the hybrid distribution consistency (HDC) to correct the knowledge deficiency of the Follower in the denoising network. Finally, to enhance the knowledge capacity of the Follower and improve its robustness to data variations, we design a time-dependent conditional encoder (TCE), distinct from the Leader's, to extract a more diverse knowledge distributions, and incorporate cross-consistency (CC) learning to reduce its sensitivity to distribution variation. In summary, our contributions can be outlined as:
\vspace{-1.5mm}
\begin{itemize}[itemsep=0.1em]
\item We present KRNet, the first method to study low-quality data COD. KRNet leverages knowledge rectification to obtain segmentation results from low-quality inputs that resemble high-quality ones, bridging the performance gap between low- and high-quality data.
\item We propose a conditional diffusion model based Leader-Follower knowledge rectification framework, where CDC and HDC strategies significantly simplify the design of knowledge rectification for low-quality COD. We further introduce TCE and CC learning to enhance the knowledge capacity of the Follower and reduce its sensitivity to distribution variations.
\item Extensive experiments on multiple datasets demonstrate that KRNet effectively rectifies the knowledge learned from low-quality data, achieving superior segmentation performance across various validation protocols and significantly outpacing existing COD methods, with performance improvements of 8.15\%, 8.53\%, and 16.16\% for metrics $E_\phi$, $S_\alpha$, $F_\beta^\omega$ on COD10K.
\end{itemize}

\section{Related Work}
{\bf Camouflaged Object Detection (COD).} Learning-based methods have emerged as the dominant approach in COD. These methods can be roughly categorized into basic and specific tasks based on their purposes. Basic tasks focus on utilizing annotated data to explore and interpret camouflage mechanisms. Early studies paved the way for this field by drawing inspiration from nature \cite{fan2020camouflaged, mei2021camouflaged, guan2024idenet}. In addition, information-assisted strategies, like assisted edge information \cite{sun2022boundary}, depth map \cite{wang2023depth}, and texture cues \cite{zhu2022can}, provide richer contextual insights that support break camouflage. Recently, specific tasks have evolved to meet the challenges that COD encounters in various real-world contexts. Weakly supervised learning methods employ scribble annotations \cite{he2023weakly} or points \cite{10.1007/978-3-031-72761-0_18}, reducing dependence on high-quality mask annotations. Moreover, the method investigates open-vocabulary COD \cite{pang2025open} or explores text-prompted segmentation \cite{huleveraging, he2024text}, revealing the prospects of multimodal learning in COD. Nevertheless, current methods rely heavily on high-quality data, overlooking information loss in low-resolution scenarios. Our KRNet presents superior camouflaged object recognition performance even in low-resolution conditions, offering a robust solution to one of the most pressing challenges in COD.

{\bf Emerging Applications of Diffusion Models.} Diffusion models \cite{ho2020denoising, dhariwal2021diffusion}, a class of generative models parameterized through Markov chains, are centered on learning a progressive denoising trajectory from a predefined noise distribution. These models have demonstrated remarkable potential across diverse modalities, including image \cite{zhang2023adding}, text \cite{zhang2023adding}, and audio \cite{huang2023make} generation. Beyond generative tasks, diffusion models have been increasingly explored in various downstream applications in computer vision, like low-level vision \cite{lugmayr2022repaint, yue2024resshift} and dense prediction tasks \cite{baranchuklabel, chen2024camodiffusion}, underscoring their broad adaptability across diverse tasks.
However, applying diffusion models to dense prediction tasks (i.e., COD) on low-quality images remains a promising area that requires further exploration.

\section{The Proposed Method}
\subsection{Preliminary}
{\bf Diffusion model} is a two-stage process comprising forward noise addition and reverse denoising \cite{ho2020denoising}. Its primary objective is to learn a denoising model based on the forward noise addition process over $T$ steps. Given observed data $\bm{x}_0\sim q(\bm{x}_0)$, the forward process of the Markov chain, $q(\bm{x}_{1:T}|\bm{x}_0):=\prod_{t=1}^{T}(\bm{x}_{t}|\bm{x}_{t-1})$, progressively transforms $\bm{x}_0$ into Gaussian noise:
\begin{equation}\label{Eq1}
q(\bm{x}_t|\bm{x}_0) = \mathcal{N}(\bm{x}_t; \sqrt{\bar{\alpha}_t} \bm{x}_0, (1 - \bar{\alpha}_t)\mathbf{I})
\end{equation}
where $\bar{\alpha}_t=\prod_{t=1}^{T}\alpha_t$, $\alpha_{t}\in (0,1]$, and $\{\alpha_{1}$, $...$, $\alpha_{t}\}$ represents a predefined, non-learnable sequence that monotonically decreases as $T$ increases.
The reverse process is a learnable denoising process parameterized by $\theta$, denoted as $p_{\theta}(\bm{x}_0)=\int p_{\theta}(\bm{x}_{0:T})\mathrm{d}\bm{x}_{1:T}$, where $p_{\theta}(\bm{x}_{0:T}):=p_{\theta}(\bm{x}_T)\prod_{t=1}^{T}p_{\theta}(\bm{x}_{t-1}|\bm{x}_t)$ is a Markov chain sampling from $\bm{x}_T$ to $\bm{x}_0$. And given $p_\theta(\bm{x}_T)=\mathcal{N}(\bm{x}_T;\mathbf{0},\mathbf{I})$, then:
\begin{equation}\label{Eq2}
p_{\theta}(\bm{x}_{t-1}|\bm{x}_t) = \mathcal{N}(\bm{x}_{t-1};\bm{\mu}_{\theta}(\bm{x}_t,t), \bm{\Sigma}_{\theta}(\bm{x}_t,t))
\end{equation}
The denoising learning process learns to match the joint distribution of the forward process by optimizing the evidence lower bound $-\mathcal{L}_{\theta}(\bm{x}_0)\leq\mathrm{log}p_\theta(\bm{x}_0)$:
 \begin{equation}
 \begin{split}
     \mathcal{L}_\theta(\bm{x}_0) &= \mathbb{E}_{q} \Big[ D_{\mathrm{KL}}(q(\bm{x}_{t}|\bm{x}_0) \parallel p_{\theta}(\bm{x}_{T})) - \log p_{\theta}(\bm{x}_0|\bm{x}_1) \\ & +\sum_{t>1} D_{\mathrm{KL}}(q(\bm{x}_{t-1}|\bm{x}_t, \bm{x}_0) \parallel p_{\theta}(\bm{x}_{t-1}|\bm{x}_t)) \Big] 
 \end{split}
 \end{equation}
According to the design of DDPM \cite{ho2020denoising}, this optimization objective is simplified to:
\begin{equation}\label{Eq4}
\arg\min_{\theta}\sum_t\mathbb{E}_q\left[ D_{\mathrm{KL}}(q(\bm{x}_{t-1}|\bm{x}_t, \bm{x}_0)\parallel p_{\theta}(\bm{x}_{t-1}|\bm{x}_t)) \right]
\end{equation}

{\bf Conditional diffusion model} \cite{saharia2022image} associates a conditioning signal $\bm{c}$ with the data $\bm{x}_0$, jointly sampled from the data distribution, $(\bm{x}_0,\bm{c})\sim q(\bm{x_0},\bm{c})$. Generally, a learnable conditional model $p_\theta(\bm{x}_0|\bm{c})$ is employed to steer the reverse denoising process. For example, CamoDiffusion \cite{chen2024camodiffusion} utilizes image features as conditional signal to generate denoising masks, revising the reverse process Eq. (\ref{Eq2}) as:
\begin{equation}
    p_{\theta}(\bm{x}_{t-1}|\bm{x}_t,\bm{c}) = \mathcal{N}(\bm{x}_{t-1};\bm{\mu}_{\theta}(\bm{x}_t,t,\bm{c}), \bm{\Sigma}_{\theta}(\bm{x}_t,t,\bm{c}))
\end{equation}
Accordingly, the optimization objective for Eq. (\ref{Eq4}) is adapted to include the conditional signal $\bm{c}$, reformulating it to:
\begin{equation}\label{Eq6}
\arg\min_{\theta}\sum_t\mathbb{E}_q\left[ D_{\mathrm{KL}}(q(\bm{x}_{t-1}|\bm{x}_t, \bm{x}_0)\parallel p_{\theta}(\bm{x}_{t-1}|\bm{x}_t,\bm{c})) \right]
\end{equation}
Notably, the conditional signals leave the forward noising process unaffected, allowing attention to focus only on their role in the reverse process. Next, we detail how conditional signal enables robust COD within low-quality data.

\subsection{Problem Setup and Overview}
This work introduces the first approach to COD for low-quality data, addressing a long-standing gap in segmentation performance between degraded inputs and high-quality outputs, a challenge previously overlooked by conventional methods. Specifically, low-quality paired data $\mathcal{D}^l=\{\bm{x}^l,\bm{m}^l\}$ with image $\bm{x}^l$ and mask $\bm{m}^l$ experiences significant information loss, leading to insufficient decision-making cues that severely impair COD performance. The key challenge lies in recovering this missing information to expand the knowledge capacity of models trained on degraded data. To overcome this, we propose using knowledge distillation \cite{hinton2015distilling}, where the distribution learned from existing high-quality data serves as a gold standard knowledge to correct the distribution learned from low-quality data, effectively bridging the knowledge gap. However, the intricate design of existing COD models complicates the implementation of distillation, making it difficult to determine where and how to apply it. We identify that conditional diffusion models offer a simplified and effective alternative which provides the possibility for optimization focusing only on $\bm{c}$, as shown in Eq. (\ref{Eq6}). 

Specifically, we establish a conditional CamoDiffusion-based \cite{chen2024camodiffusion} Leader-Follower framework, as illustrated in Figure \ref{Fig:framework}, where both models are equipped with a conditional encoder $E$ and a denoising network $Z$. Within the Leader, given high-quality data $\mathcal{D}^h=\{\bm{x}^h,\bm{m}^h\}$, the model aim to get clean mask $\bm{m}_0^h$ conditions on $\bm{c}^h=E_h(\bm{x}^h, \phi_h)$ to train the denoising network $Z_h:=p_\theta(\bm{m}_0^h|\bm{m}_t^h, t, \bm{c}^h)$ derived from Eq. (\ref{Eq6}) and conditional encoder $E_h$, applying reverse processes on the noise mask $\bm{m}_t^h$. It enables the generation of high-quality segmentation masks during the sampling phase, utilizing only $\bm{x}^h$ and noise $\bm{\epsilon}$. Since the sampling phase is mask-agnostic, we simplify our framework by designing a training process where the Leader and Follower share the same input mask $\bm{m}^h$, and extend same training approach to low-quality data by training $Z_l := \hat{p}_\theta(\hat{\bm{m}}_0^h|\bm{m}_{t^\prime}^h, {t^\prime}, \bm{c}^l)$ and $E_l$, with the performance gap arising from the conditional signal $\bm{c}^l$. 

{\bf Assumption 1:} {\it If $\bm{c}^l$ aligns with the gold-standard $\bm{c}^h$, the model $\hat{p}_\theta$ can achieve high-quality segmentation like $p_\theta$, and we arrive at the following formula:}
\begin{equation}\label{Eq7}
\begin{split}
\mathbb{E}_{\bm{c}^l \sim p(\bm{c}^l)} \left[ Z_l(\bm{m}_{t^\prime}^h, t^\prime, \bm{c}^l) \right] 
&\approx 
\mathbb{E}_{\bm{c}^h \sim p(\bm{c}^h)} \left[ Z_h(\bm{m}_t^h, t, \bm{c}^h) \right],\\
\text{if} \quad p(\bm{c}^l)&\approx p(\bm{c}^h).
\end{split}
\end{equation}
Building upon the analysis and Assumption 1, we reframe the high-quality distribution-guided knowledge distillation process as a CDC learning within a diffusion model.

\subsection{Knowledge Rectification with Distribution Consistency}
Following the assumption proposed in Eq. (\ref{Eq7}), bridging the performance gap between low-quality and high-quality data relies on correcting the low-quality conditional distribution $\bm{c}^l$ using the gold-standard conditional distribution $\bm{c}^h$. However, as Figure \ref{Fig:framework} illustrates, 
$\bm{c}^l$ also contributes to the computational flow of the denoising decoder. Therefore, we decompose the knowledge rectification objective for distribution consistency into two subproblems: conditional and hybrid distribution consistency.

{\bf Conditional distribution consistency} (CDC) aims to perform an initial rectification of the $\bm{c}^l$ before injecting it into the denoising network. Specifically, leveraging Eq. (\ref{Eq7}), we derive the first optimization objective as follows:
\begin{equation}\label{Eq8}
 \min_{\phi_l} D_{\text{dis}}(E_l(\bm{x}^l,\phi_l), E_h(\bm{x}^h,\phi_h) \big|_{\nabla=0})
\end{equation}
This objective utilizes a conditional encoder $E^\alpha, \alpha \in\{l,h\}$ to extract conditional distributions $\bm{c}^\alpha$ from both low- and high-quality data. A distribution metric function $D_\text{{dis}}$ is then employed to rectify $\bm{c}^l$, enabling $E^l$ to attain knowledge that mimics high-quality distributions. During training, we optimize only the parameters of $\phi _l$, keeping $\phi_h$ frozen to preserve the stability of the gold standard knowledge.

\begin{figure*}[t]
	\centering
	\includegraphics[width=1.0\linewidth]{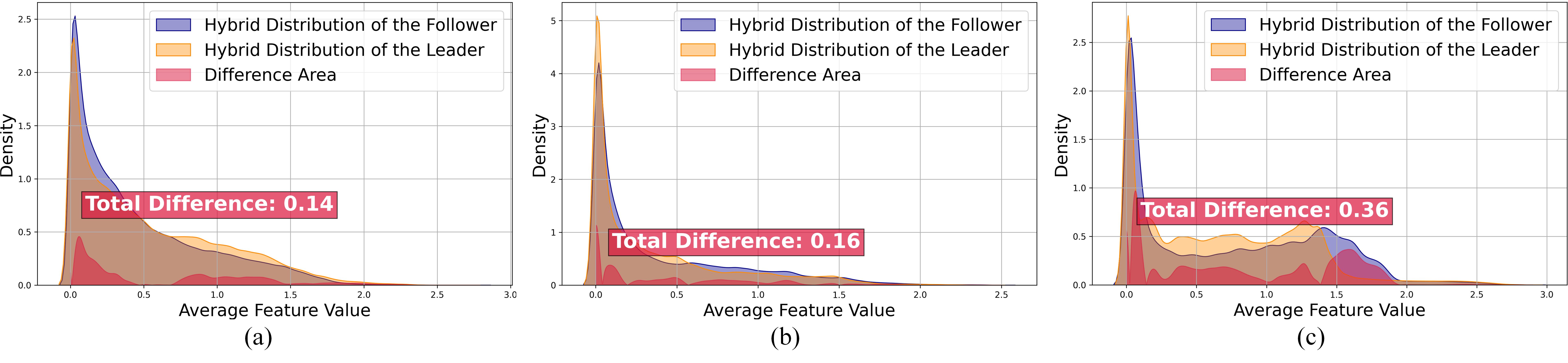}
        \vspace{-20pt}
	\caption{The differences in the hybrid distributions between the Leader and Follower. (a)-(c) represent the differences in hybrid distributions across the first to third layers of the denoising decoder. The gradually increasing discrepancies in the hybrid distributions make it challenging for the Follower to achieve high-quality segmentation results. The introduction of HDC can further rectify these knowledge discrepancies.}
	\label{Fig:hy_dist}
    \vspace{-1.5mm}
\end{figure*}
{\bf Hybrid distribution consistency} (HDC) strives to correct the knowledge embedded in the hybrid distribution formed by the noise and conditional distributions within the denoising decoder $Z^D$. This correction manages two critical challenges that initial knowledge rectification fails to resolve during the denoising phase: (1) the growing discrepancy of the conditional distribution, and (2) the divergence of the noise distributions. The discrepancy in conditional distribution is intrinsic, as Eq. (\ref{Eq8}) yields an approximate solution. We find that after $\bm{c}^l$ is injected into the denoising network, the distributional discrepancies exhibit a progressively amplifying trend, as demonstrated in Figure \ref{Fig:hy_dist}. Furthermore, during model optimization, the Follower and Leader models are provided with distinct noise input data $\bm{m}_{t^{\prime}}^h$ and $\bm{m}_t^h$, respectively, assigned to different timesteps, to enhance the robustness of the Follower. Consequently, the hybrid distribution discrepancy stemming from the divergence of the noise distributions also requires further rectification. Therefore, the extraction of the decoder module $Z_\alpha^D, \alpha \in\{l,h\}$ with hybrid distribution $\bm{d}^\alpha$ from Eq. (\ref{Eq7}) further derives the knowledge rectification expressed in the Assumption 2.

{\bf Assumption 2:} {\it When $\bm{d}^l$ is calibrated using $\bm{d}^h$ as the gold-standard distribution, $\hat{p}_\theta$ can produce higher-quality segmentation that more closely matches $p_\theta$:}
\begin{equation}
\begin{split}
\mathbb{E}_{\bm{d}^l \sim p(\bm{d}^l)} \left[ Z_l^D(\bm{d}^l, t^\prime)\right] 
&\approx 
\mathbb{E}_{\bm{d}^h \sim p(\bm{d}^h)} \left[ Z_h^D(\bm{d}^h, t)\right],\\
\text{if} \quad p(\bm{d}^l)&\approx p(\bm{d}^h).
\end{split}
\end{equation}
Thus, the second optimization objective can be obtained:
\begin{equation}\label{Eq10}
\min_{\theta^D_l} D_{\text{dis}}\left(Z_l^D(\bm{d}^l, \theta_l^D), Z_h^D(\bm{d}^h, \theta^D_h) \big|_{\nabla=0}\right)
\end{equation}
Similarly, we optimize only the parameters $\theta_l^D$ of Follower, while keeping the Leader's parameters frozen.

\subsection{Knowledge Rectification with Cross-consistency}\label{Sec:corss}
In COD, handling data with varying resolutions and diverse scenes \cite{guan2024idenet} presents significant challenges for knowledge rectification, often leading to instability, particularly under different architectures. Inspired by contrastive learning \cite{grill2020bootstrap}, we propose a cross-consistency (CC) knowledge rectification strategy to optimize the objectives defined in Eq. (\ref{Eq8}) and Eq. (\ref{Eq10}), promoting the learning of generalized knowledge robust to perturbations. Specifically, as illustrated in Figure \ref{Fig:framework}, we introduce paired different augmented views ($\bm{x}_a^l,\bm{x}_A^l$) of the same image as inputs to the Follower, extracting their conditional distribution ($\bm{c}_a^h,\bm{c}_A^h$) and the hybrid distribution ($\bm{d}_a^l,\bm{d}_A^l$). Similarly, the Leader processes ($\bm{x}_a^h,\bm{x}_A^h$) to obtain their conditional distributions ($\bm{c}_a^h,\bm{c}_A^h$) and hybrid distribution for ($\bm{d}_a^h,\bm{d}_A^h$). Hence, the optimization objectives are updated as follows:
\begin{equation}\label{Eq11}
\left\{
\begin{array}{l}
  \min\limits_{\phi_l} \sum\limits_{\beta} D_{\text{dis}}(E_l(\bm{x}_\beta^l,\phi_l), E_h(\bm{x}_\beta^h,\phi_h) \big|_{\nabla=0}) \\[8pt]
  \min\limits_{\theta^D_l} \sum\limits_{\beta} D_{\text{dis}}(Z_l^D(\bm{d}_\beta^l, \theta_l^D), Z_h^D(\bm{d}_\beta^h, \theta^D_h) \big|_{\nabla=0})
\end{array}
\right.
\end{equation}
where $\beta \in\{a,A\}$. This strategy explicitly models the distributional relationships between different augmented views and effectively captures feature consistency and variation across these views. Consequently, it further prompts the learning of stable knowledge representations for the Follower, even under conditions of data variability.
\begin{table*}[htp]
	\vspace{-2mm}
	\centering
	\caption{The quantitative comparison between our KRNet and SOTA methods. EDSR+COD refers to the results obtained by first applying super-resolution on low-quality data and then performing COD. All results are based on low-quality data with 4$\times$ downsampling. $``\uparrow"$ and $``\downarrow"$ represent that the higher and the lower the better respectively. \colorbox{cyan!30}{\textbf{Bold}} is the best result.}
	\renewcommand{\arraystretch}{0.8}
	\setlength{\tabcolsep}{3.8pt}{
		
		\begin{tabular}{c|c|cccc|cccc|cccc}
			\toprule
			\multirow{2}{*}{\textbf{Methods}}&\multirow{2}{*}{\textbf{Publication}}&\multicolumn{4}{c}{CAMO}& \multicolumn{4}{|c|}{COD10K} &\multicolumn{4}{c}{NC4K}\\
			\cmidrule{3-6}
			\cmidrule{7-10}
			\cmidrule{11-14}
			&& $S_\alpha$$\uparrow$& $E_\phi$$\uparrow$& $F_{\beta}^{\omega}$$\uparrow$& $M$$\downarrow$& $S_\alpha$$\uparrow$& $E_\phi$$\uparrow$& $F_{\beta}^{\omega}$$\uparrow$& $M$$\downarrow$&$S_\alpha$$\uparrow$& $E_\phi$$\uparrow$& $F_{\beta}^{\omega}$$\uparrow$& $M$$\downarrow$\\
			\midrule
        \multicolumn{14}{c}{COD}\\
        \midrule 
        SINet&CVPR2020&65.05&67.83&46.66&13.10&71.81&76.40&48.56&5.97&73.94&77.81&58.00&8.64\\
        PFNet&CVPR2021&69.83&77.03&55.60&11.64&72.98&81.56&53.40&5.73&75.35&81.99&61.61&8.14\\
        SegMaR&CVPR2022&39.80&29.92&1.90&19.43&44.74&29.50&0.41&9.84&41.30&29.58&0.81&16.45\\
        BGNet&IJCAI2022&68.95&76.13&55.65&11.46&74.43&83.14&57.18&5.11&75.89&83.04&64.08&7.58\\
        FSNet&TIP2023&38.56&34.06&2.50&20.95&43.21&38.65&1.81&12.67&39.97&35.32&2.38&18.55\\
        HitNet&AAAI2023&70.41&77.69&58.56&10.51&75.34&83.47&60.23&4.52&77.40&84.07&67.52&6.66\\
        IdeNet&TIP2024&	72.70&79.66&60.44&10.65&	77.03&85.56&61.89&4.52&	79.72&86.29&70.03&6.32\\
        RISNet&CVPR2024&40.00&28.09&0.19&18.50&44.65&34.51&0.07&9.43&	41.48&30.14&0.12&15.65\\
    CamoDiffusion&AAAI2024&39.94&29.55&0.71&18.95&44.54&33.36&0.33&10.34&41.23&31.46&0.36&16.36\\
        \midrule
        \multicolumn{14}{c}{EDSR+COD}\\
        \midrule
        SINet&CVPR2020&62.45&56.31&32.20&20.05&61.00&57.27&22.87&14.4&65.16&59.25&33.00&16.68\\
        PFNet&CVPR2021&69.15&77.88&55.37&11.67&73.11&82.05&54.35&5.57&75.46&82.80&62.92&7.83\\
        BGNet&IJCAI2022&66.05&71.59&51.65&11.95&74.00&81.25&57.76&4.74&75.67&81.84&64.93&7.25\\
        HitNet&AAAI2023&72.30&80.84&61.14&10.54&76.58&85.49&61.72&4.51&79.02&86.44&69.48&6.40\\
        CamoDiffusion&AAAI2024&77.85&86.01&70.08&8.26&81.01&88.93&69.52&3.69&82.11&88.90&74.80&5.37\\
       \midrule
        \midrule
KRNet(Ours)&-&\cellcolor{cyan!30}\textbf{83.49}&\cellcolor{cyan!30}\textbf{90.67}&\cellcolor{cyan!30}\textbf{78.96}&\cellcolor{cyan!30}\textbf{5.50}&\cellcolor{cyan!30}\textbf{85.56}&\cellcolor{cyan!30}\textbf{93.71}&\cellcolor{cyan!30}\textbf{78.05}&\cellcolor{cyan!30}\textbf{2.13}&\cellcolor{cyan!30}\textbf{86.92}&\cellcolor{cyan!30}\textbf{93.72}&\cellcolor{cyan!30}\textbf{82.89}&\cellcolor{cyan!30}\textbf{3.15}\\
			\bottomrule
	\end{tabular}}\label{tab:main}
	\vspace{-1.5mm}
\end{table*}
\subsection{Time-dependent Conditional Encoder Fine-tuning}
Different model architectures capture data representations at various levels of abstraction. By using distinct architectures for knowledge distillation, we can enhance feature diversity and implicitly increase the knowledge capacity of the Follower, improving COD performance under low-quality data. To achieve this, we employ Dinov2 \cite{oquab2023dinov2} as the time-dependent conditional encoder (TCE) for the Follower, distinct from the Leader's. We implement time tokens $\bm{t}_n$ at each layer of Dinov2 to fine-tune its features $\bm{f}_n$ concerning temporal dependencies, strengthening its guidance in the denoising process:
\begin{equation}
\bm{f}_{n} = \text{Layer}_n\big(\text{Concat}(\bm{f}_{n-1}, \text{Embedding}(\bm{t}_n))\big)
\end{equation}
where $\bm{f}_0 = \bm{f}_{\text{init}}, n \in \{1, 2, \dots, 24\}$. We then fuse features from four layers of Dinov2 to generate $\bm{c}^l$, as done with the conditional encoder in the Leader:
\begin{equation}
\bm{c}^l = \text{Fuse}(\bm{f}_7, \bm{f}_{11}, \bm{f}_{15}, \bm{f}_{23})
\end{equation}

\section{Experiments}
\subsection{Experimental Settings}
{\bf Datasets.} We validate our method for low-quality COD on three widely used datasets: CAMO \cite{le2019anabranch}, COD10K \cite{fan2020camouflaged}, and NC4K \cite{lv2021simultaneously}. For training, we use a subset of CAMO and COD10K, comprising 4,040 images as the training set. Low-quality data at scales of 2×, 4×, and 8× are generated using bicubic downsampling, a common approach in image super-resolution \cite{lim2017enhanced}. More details about the datasets are provided in Appendix.

{\bf Implementation Details.} We implement KRNet using PyTorch and conduct training and inference on a single NVIDIA H100 GPU with 80GB of memory. During training, the Leader loads pre-trained parameters optimized on high-quality data and remains frozen, while the Follower is trained from scratch using the extracted distribution of Leader as a reference. The training process employs a batch size of 20, an initial learning rate of 1e-4, and a total of 150 epochs with images resized to 384$\times$384. We optimize the segmentation performance using the AdamW \cite{loshchilov2017decoupled} optimizer and Structure Loss \cite{wei2020f3net}. Further details can be found in Appendix.

{\bf Evaluation protocols.} Following previous COD methods, we evaluate performance using four commonly adopted metrics: S-measure $S_\alpha$ \cite{fan2017structure}, mean E-measure $E_\phi$ \cite{10.5555/3304415.3304515}, weighted F-measure $F_\beta^\omega$ \cite{margolin2014evaluate}, and mean absolute error $M$ \cite{perazzi2012saliency}. 

\subsection{Comparison with SOTA Methods}\label{sec:4.2}
We conduct a comprehensive analysis of our proposed KRNet in comparison with 9 SOTA COD methods, including SINet \cite{fan2020camouflaged}, PFNet \cite{mei2021camouflaged}, SegMaR \cite{jia2022segment}, BGNet \cite{sun2022boundary}, FSNet \cite{song2023fsnet}, HitNet \cite{hu2023high}, IdeNet \cite{guan2024idenet}, RISNet \cite{wang2024depth}, and CamoDiffusion \cite{chen2024camodiffusion}, in terms of both quantitative and qualitative performance. We also evaluate the performance of the top-performing methods of each year with the image super-resolution method EDSR \cite{lim2017enhanced}. 

\begin{figure*}[t]
	\centering
		\includegraphics[width=\linewidth]{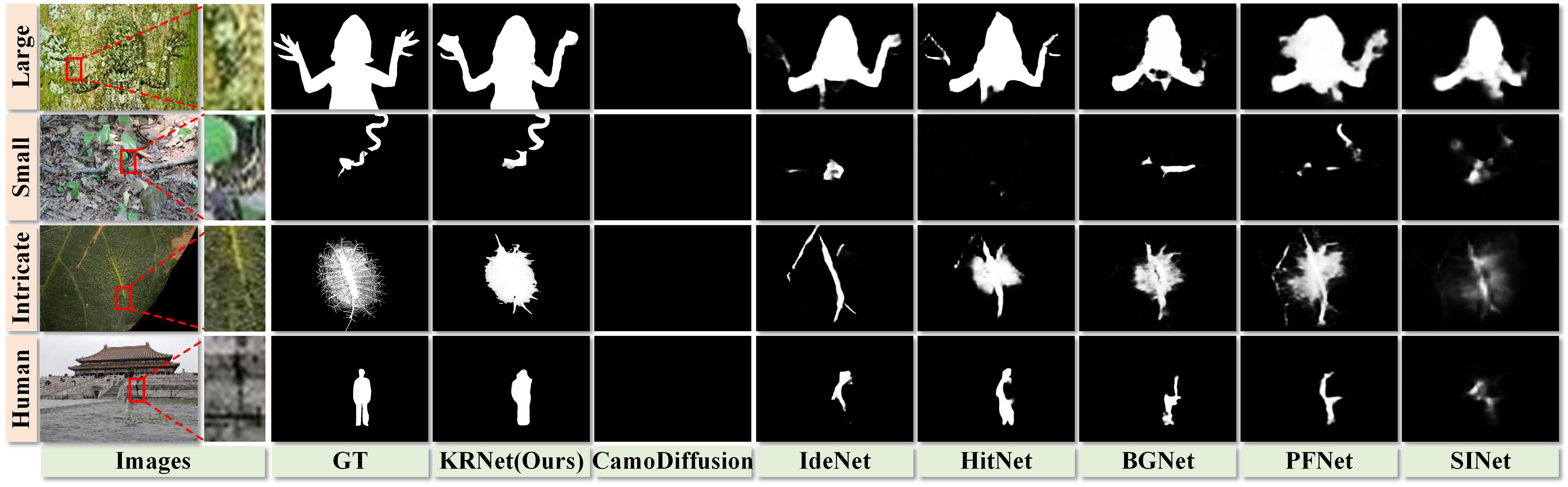}
    \vspace{-6mm}
	\caption{Visual Quality Comparison with SOTA Methods. The figure presents various common target types, including large, small, intricate details, and humans objects. The magnified patch details reveal that crucial target information is significantly missing in low-quality images. More results are shown in Appendix.}
	\label{Fig:comp}
    \vspace{-2mm}
\end{figure*}
\begin{table*}[!ht]
\vspace{-1mm}
  \begin{minipage}{0.6\linewidth} 
    \centering
	\caption{Comparison of quantization results across different models. \colorbox{green!30}{\textbf{Bold}} represents the best results with the same conditional encoder, while \colorbox{cyan!30}{\textbf{Bold}} shows the best results with the introduction of TCE.}
	\renewcommand{\arraystretch}{0.8}
	\setlength{\tabcolsep}{2pt}{
	\begin{tabular}{cccc|cccc|cccc}
			\toprule
			\multicolumn{4}{c}{Methods}& \multicolumn{4}{|c|}{CAMO} &\multicolumn{4}{c}{NC4K}\\
			\cmidrule{1-4}
			\cmidrule{5-8}
			\cmidrule{9-12}
			CDC& HDC& CC&TCE& $S_\alpha$$\uparrow$& $E_\phi$$\uparrow$& $F_{\beta}^{\omega}$$\uparrow$& $M$$\downarrow$& $S_\alpha$$\uparrow$& $E_\phi$$\uparrow$& $F_{\beta}^{\omega}$$\uparrow$& $M$$\downarrow$\\
			\midrule
                \ding{55}&	\ding{55}&	\ding{55}&\ding{55}&39.94&29.55&0.71&18.95&41.23&31.46&0.36&16.36	\\
                 \ding{51}&	\ding{55}&	\ding{55}&\ding{55}&78.33&86.76&71.59&7.93&82.96&89.85&76.43&4.80\\
                  \ding{55}&	\ding{51}&	\ding{55}&\ding{55}&77.14&85.55&69.76&8.15&82.86&89.88&76.21&4.84\\
               \ding{51}&	\ding{51}&	\ding{55}&\ding{55}&79.02&87.77&72.63&7.46&83.26&90.09&76.75&4.73\\
               \ding{51}&	\ding{51}&	\ding{51}&\ding{55}&\cellcolor{green!30}\textbf{79.41}&\cellcolor{green!30}\textbf{87.77}&\cellcolor{green!30}\textbf{72.88}&\cellcolor{green!30}\textbf{7.32}&\cellcolor{green!30}\textbf{83.36}&\cellcolor{green!30}\textbf{90.11}&\cellcolor{green!30}\textbf{77.05}&\cellcolor{green!30}\textbf{4.68}\\
               \ding{55}&	\ding{55}&	\ding{55}&\ding{51}&40.96&25.06&0.12&18.12&42.48&25.12&0.07&15.28\\
                \ding{55}&	\ding{51}&	\ding{55}&\ding{51}&82.35&89.38&77.07&5.85&86.88&\cellcolor{cyan!30}\textbf{93.83}&82.40&3.22\\
                 \ding{51}&	\ding{55}&	\ding{55}&\ding{51}&82.79&90.78&77.62&5.91&86.64&93.66&81.92&3.30\\
                  \ding{51}&	\ding{51}&	\ding{55}&\ding{51}&83.19&90.30&77.79&5.74&86.75&93.78&82.03&3.28\\
                \ding{51}&	\ding{51}&	\ding{51}&\ding{51}&\cellcolor{cyan!30}\textbf{83.49}&\cellcolor{cyan!30}\textbf{90.67}&\cellcolor{cyan!30}\textbf{78.96}&\cellcolor{cyan!30}\textbf{5.50}&\cellcolor{cyan!30}\textbf{86.92}&93.72&\cellcolor{cyan!30}\textbf{82.89}&\cellcolor{cyan!30}\textbf{3.15}\\
			\bottomrule
	\end{tabular}}\label{Tab:component_abl}
  \end{minipage}%
  \hspace{10mm}
  \begin{minipage}{0.3\linewidth} 
    \centering
    \vspace{2mm}
\caption[Short title]{\raggedright Comparison of applying HDC at different layers. \ding{172}-\ding{174} represent the first to the third layers of the denoising decoder.}
\renewcommand{\arraystretch}{1.23}
\setlength{\tabcolsep}{3pt}{
\begin{tabular}{cccccc}
\toprule
\multirow{2}{*}{\textbf{Layers}} & \multicolumn{4}{c}{\textbf{COD10K}} \\
\cmidrule(lr){2-5} 
 & $S_\alpha$$\uparrow$& $E_\phi$$\uparrow$& $F_{\beta}^{\omega}$$\uparrow$& $M$$\downarrow$ \\
\midrule
\ding{172}& 82.45&\cellcolor{cyan!30}\textbf{90.85}& 72.28& 3.07 \\
\ding{173}&82.20&90.61&71.66&3.16 \\
\ding{174} & 82.21&90.75&71.87&3.09\\
\ding{172}+\ding{174} &82.55&90.82&72.28&3.10\\
\ding{172}+\ding{173}+\ding{174}&82.33&90.48&71.93&3.12\\
\ding{173}+\ding{174}&\cellcolor{cyan!30}\textbf{82.58}&90.71&\cellcolor{cyan!30}\textbf{72.49}&\cellcolor{cyan!30}\textbf{3.04} \\
\bottomrule
\end{tabular}}
\label{Tab:layer_abl}
  \end{minipage}
  \vspace{-2mm}
\end{table*}
{\bf Quantitative Comparison.} Table \ref{tab:main} presents the quantitative results of all methods on low-quality COD data (additional results are available in Appendix). The findings reveal that SOTA COD methods face challenges in extracting sufficient camouflage-breaking cues from low-quality data, leading to notable performance degradation, particularly in the $F_\beta^\omega$, where methods such as RISNet and CamoDiffusion approach near-zero values. Integrating image super-resolution as a preprocessing step for detail restoration improves the performance of iterative refinement methods, such as CamoDiffusion and HitNet. However, methods dependent on fine details or edge information, including SINet and BGNet, exhibit a decline in performance. These suggest that while super-resolution can supplement camouflage-breaking cues, its smoothing effects (as shown in Appendix) hinder methods that rely on fine-detail compensation, thereby limiting improvements in object detection.
In contrast, our KRNet, constructed upon the CamoDiffusion baseline, outperforms all SOTA methods, even when combined with super-resolution techniques. Compared to CamoDiffusion, KRNet achieves significant improvements of 43.55\%, 41.02\%, and 45.69\% in the $S_\alpha$ across three datasets. These results substantiate that knowledge rectification provides a simple yet more effective approach on low-quality data.

{\bf Qualitative Comparison.} Figure \ref{Fig:comp} presents a visual quality comparison between KRNet and top-performing SOTA methods (more results are shown in Appendix). The results demonstrate that KRNet provides more accurate and complete target segmentation across diverse scenarios, including large targets, small targets, intricate details, and human subjects. In contrast, SOTA methods, constrained by limited knowledge, struggle to capture all relevant information about the targets. Specifically, many of these methods fail to detect camouflaged targets, particularly in scenes with small targets or intricate details. These findings underscore the importance of knowledge rectification, which can push beyond existing knowledge boundaries and uncover more valuable camouflage cues.

\subsection{Ablation Studies}
{\bf Contribution Analysis of Model Components.} We systematically evaluate the contributions of KRNet components by adding or removing them. As shown in Table \ref{Tab:component_abl}, the first five rows indicate that incrementally introducing CDC, HDC, and CC leads to notable improvements in COD results under the same conditional encoder. Furthermore, introducing the TCE distinct from the Leader’s resulted in improved performance on $S_\alpha$ and $M$, while unexpectedly leading to a decline in $E_\phi$ and $F_\beta^\omega$. As we progressively incorporated our proposed methods, the performance presents a substantial boost, for instance, $S_\alpha$ improved by 42.53\% on CAMO and 44.64\% on COD10K. These findings comprehensively validate the effectiveness of our proposed components for low-quality COD, regardless of whether the Leader and Follower employ identical or distinct conditional encoders.

\begin{table}[h!]
\centering
\vspace{-2mm}
\caption{Comparison of different loss functions on CDC and HDC. The experiment is validated on COD10K.}
\setlength{\tabcolsep}{1.6pt}{%
\begin{tabular}{c|c|c|c|c|c|c|c|c}
\toprule
\multirow{2}{*}{\textbf{Loss}} & \multicolumn{4}{c}{\textbf{CDC}} & \multicolumn{4}{c}{\textbf{HDC}} \\
\cmidrule(lr){2-5} \cmidrule(lr){6-9}
 & $S_\alpha$$\uparrow$& $E_\phi$$\uparrow$& $F_{\beta}^{\omega}$$\uparrow$& $M$$\downarrow$ & $S_\alpha$$\uparrow$& $E_\phi$$\uparrow$& $F_{\beta}^{\omega}$$\uparrow$& $M$$\downarrow$ \\
\midrule
MAE & 81.81&90.33& 71.17& 3.24 & 82.01&90.47&71.48&\cellcolor{cyan!30}\textbf{3.09}\\
MSE &80.68&89.67&68.90&3.45 & 81.83&90.26&71.19&3.20 \\
MMD & 81.68&90.18&70.94&3.19&81.38&89.89&70.46&3.34\\
FA&80.63&89.38&68.84&3.48&81.10&89.52&69.79&3.30 \\
CS &81.37&89.46&70.05&3.38&81.32&89.58&70.35&3.41\\
KL&\cellcolor{cyan!30}\textbf{82.21}&\cellcolor{cyan!30}\textbf{90.75}&\cellcolor{cyan!30}\textbf{71.87}&\cellcolor{cyan!30}\textbf{3.09}&\cellcolor{cyan!30}\textbf{82.07}&\cellcolor{cyan!30}\textbf{90.48}&\cellcolor{cyan!30}\textbf{71.61}&3.17\\
\bottomrule
\end{tabular}}
\vspace{-2mm}
\label{Tab:loss_abl}
\end{table}
\begin{table}[h!]
\centering
\caption{Comparison of fine-tuning strategies using time tokens. Implementation without CDC and with HDC on first layer and third layer.}
\renewcommand{\arraystretch}{0.8}
	\setlength{\tabcolsep}{12.8pt}{
\begin{tabular}{cccccc}
\toprule
\multirow{2}{*}{\textbf{TCE}} & \multicolumn{4}{c}{\textbf{COD10K}} \\
\cmidrule(lr){2-5} 
 & $S_\alpha$$\uparrow$& $E_\phi$$\uparrow$& $F_{\beta}^{\omega}$$\uparrow$& $M$$\downarrow$ \\
\midrule
FL& 84.85&93.07& 76.44& 2.25 \\
OL&85.06&93.40&76.81&\cellcolor{cyan!30}\textbf{2.23} \\
GL& 85.24&93.52&76.94&2.24\\
EL&\cellcolor{cyan!30}\textbf{85.30}&\cellcolor{cyan!30}\textbf{93.74}&\cellcolor{cyan!30}\textbf{77.21}&\cellcolor{cyan!30}\textbf{2.23}\\
\bottomrule
\end{tabular}}
\label{Tab:fine_abl}
\vspace{-2mm}
\end{table}
{\bf Analysis of HDC on Different Layers.} In practice, the denoising decoder primarily consists of three layers of multi-scale features. As presented in Table \ref{Tab:layer_abl}, these features illustrate the role of HDC in knowledge rectification across different layers. The results reveal distinct benefits between single-layer and multi-layer. When using a single layer, applying the constraint to the first layer (\ding{172}) of the decoder proves most effective, inducing early-stage knowledge rectification. For multi-layer consistency, applying them to features closer to the output (\ding{173}+\ding{174}) yields better performance, leading to late-stage knowledge rectification. Besides, this finding further supports the idea that correcting knowledge at multiple levels is a more effective strategy.

{\bf Effectiveness of Different Loss Functions.} Table \ref{Tab:loss_abl} investigates the effects of various distribution consistency constraints on CDC and HDC, employing several commonly used distillation losses: Mean Absolute Error (MAE), Mean Squared Error (MSE), Maximum Mean Discrepancy (MMD) \cite{ouyang2021maximum}, Feature Affinity (FA) \cite{wang2020dual}, Cosine Similarity (CS), and Kullback-Leibler Divergence (KL). The analysis leverages the same conditional encoder for the Leader and Follower. The results reveal that KL loss surpasses the others in effectively rectifying knowledge learned from low-quality data for both CDC and HDC. Meanwhile, other loss functions also contribute to refining the knowledge rectification, demonstrating that Assumption 1 and 2 are reliable and effective strategies for COD under low-quality data.

{\bf Impact of Fine-tuning Strategies with Time Tokens.} Table \ref{Tab:fine_abl} summarizes the effects of our proposed fine-tuning strategies across different layers: the first layer (FL), output layers (OL), grouped layers (GL), and every layer (EL). The detailed configurations of these strategies are elaborated in Appendix. The results demonstrate that fine-tuning an increasing number of layers, from FL to EL, consistently enhances performance. Moreover, comparing OL and GL, expanding the number of layers influenced by time tokens yields further improvements. These underscore the importance of applying time tokens to more layers and extending the effect of each time token to capture high-quality conditional distributions, thereby boosting the denoising process.

\begin{table}[t]
\centering
\vspace{-2mm}
\caption{Performance comparison under different degradation scales. '\textbf{n}$\times$' represents downsampling '\textbf{n}' times.}
\renewcommand{\arraystretch}{0.8}
	\setlength{\tabcolsep}{4.8pt}{
\begin{tabular}{ccccccc}
\toprule
\multirow{1}{*}{\textbf{Down}}&\multirow{2}{*}{\textbf{Methods}} & \multicolumn{4}{c}{\textbf{COD10K}} \\
\cmidrule(lr){3-6} 
\textbf{Scale}& & $S_\alpha$$\uparrow$& $E_\phi$$\uparrow$& $F_{\beta}^{\omega}$$\uparrow$& $M$$\downarrow$ \\
\midrule
\multirow{6}{*}{\textbf{2$\times$}}&SINet&76.03&79.97&54.35&5.22\\
&PFNet& 75.08&83.67&57.92 &4.96\\
&BGNet&79.55&86.19&67.22&3.69\\
&HitNet&81.00&88.79&69.46& 3.51\\
&CamoDiffusion&42.26&57.86&2.62&12.26\\
&KRNet(Ours)&\cellcolor{cyan!30}\textbf{88.64}&\cellcolor{cyan!30}\textbf{95.54}&\cellcolor{cyan!30}\textbf{82.81}&\cellcolor{cyan!30}\textbf{1.73}\\
\midrule
\multirow{6}{*}{\textbf{8$\times$}}&SINet&64.65&69.15&36.36&7.71\\
&PFNet&66.06&74.94&41.52&7.89\\
&BGNet&62.78&68.91&38.09&8.27\\
&HitNet&68.67&79.47&48.68&6.52\\
&CamoDiffusion&45.78&25.65&0.03&9.14\\
&KRNet(Ours)&\cellcolor{cyan!30}\textbf{77.89}&\cellcolor{cyan!30}\textbf{87.06}&\cellcolor{cyan!30}\textbf{65.15}&\cellcolor{cyan!30}\textbf{3.47}\\
\bottomrule
\end{tabular}}
\label{Tab:scale_abl}
\vspace{-2mm}
\end{table}
{\bf Results with Different Scale Factors.} 
We analyzed 2$\times$ and 8$\times$ bicubic downsampling, commonly used in image super-resolution. As shown in Table \ref{Tab:scale_abl}, we compared the top-1 method of each year. Results show that 8$\times$ downsampling significantly degrades the performance of these methods, highlighting their reliance on high-quality data. Moreover, different methods demonstrate unstable performance across varying quality levels, with BGNet failing to consistently outperform PFNet and CamoDiffusion exhibiting noticeable fluctuations. However, the proposed KRNet equipped with knowledge rectification, demonstrates superior stability and performance, successfully generalizing to diverse degradation.

\section{Conclusion}
In this work, we propose KRNet, the first method for low-quality data COD, which formulates knowledge rectification to achieve high-quality-like performance on low-quality inputs. Specifically, we analyze and identify the conditional diffusion model as a simple yet effective foundation for knowledge rectification. Building on this insight, KRNet constructs a Leader-Follower framework based on several key assumptions. The frozen Leader extracts conditional and hybrid distributions as gold-standard references, enabling the Follower to refine the knowledge learned from low-quality data. Moreover, we introduce a time-dependent conditional encoder and cross-consistency to consistently boost the knowledge capacity and correct inherent knowledge biases of the Follower. Extensive comparative experiments validate our assumptions and demonstrate the prospect of KRNet for low-quality data COD. Our future work will extend KRNet to a broader range of degraded data scenarios for generalized COD.

\section*{Impact Statement}
 This paper presents work whose goal is to advance the field
 of Machine Learning. There are many potential societal
 consequences of our work, none which we feel must be
 specifically highlighted here.
 
\nocite{langley00}

\bibliography{example_paper}

\begin{thebibliography}{38}
\providecommand{\natexlab}[1]{#1}
\providecommand{\url}[1]{\texttt{#1}}
\expandafter\ifx\csname urlstyle\endcsname\relax
  \providecommand{\doi}[1]{doi: #1}\else
  \providecommand{\doi}{doi: \begingroup \urlstyle{rm}\Url}\fi

\bibitem[Baranchuk et~al.()Baranchuk, Voynov, Rubachev, Khrulkov, and Babenko]{baranchuklabel}
Baranchuk, D., Voynov, A., Rubachev, I., Khrulkov, V., and Babenko, A.
\newblock Label-efficient semantic segmentation with diffusion models.
\newblock In \emph{International Conference on Learning Representations}.

\bibitem[Chen et~al.(2025)Chen, Wei, Guo, and Gao]{10.1007/978-3-031-72761-0_18}
Chen, H., Wei, P., Guo, G., and Gao, S.
\newblock Sam-cod: Sam-guided unified framework for weakly-supervised camouflaged object detection.
\newblock In \emph{European Conference on Computer Vision}, pp.\  315--331. Springer, 2025.

\bibitem[Chen et~al.(2024)Chen, Sun, and Lin]{chen2024camodiffusion}
Chen, Z., Sun, K., and Lin, X.
\newblock Camodiffusion: Camouflaged object detection via conditional diffusion models.
\newblock In \emph{Proceedings of the AAAI Conference on Artificial Intelligence}, volume~38, pp.\  1272--1280, 2024.

\bibitem[Dhariwal \& Nichol(2021)Dhariwal and Nichol]{dhariwal2021diffusion}
Dhariwal, P. and Nichol, A.
\newblock Diffusion models beat gans on image synthesis.
\newblock \emph{Advances in neural information processing systems}, 34:\penalty0 8780--8794, 2021.

\bibitem[Fan et~al.(2017)Fan, Cheng, Liu, Li, and Borji]{fan2017structure}
Fan, D.-P., Cheng, M.-M., Liu, Y., Li, T., and Borji, A.
\newblock Structure-measure: A new way to evaluate foreground maps.
\newblock In \emph{Proceedings of the IEEE international conference on computer vision}, pp.\  4548--4557, 2017.

\bibitem[Fan et~al.(2018)Fan, Gong, Cao, Ren, Cheng, and Borji]{10.5555/3304415.3304515}
Fan, D.-P., Gong, C., Cao, Y., Ren, B., Cheng, M.-M., and Borji, A.
\newblock Enhanced-alignment measure for binary foreground map evaluation.
\newblock In \emph{Proceedings of the 27th International Joint Conference on Artificial Intelligence}, pp.\  698–704. AAAI Press, 2018.
\newblock ISBN 9780999241127.

\bibitem[Fan et~al.(2020)Fan, Ji, Sun, Cheng, Shen, and Shao]{fan2020camouflaged}
Fan, D.-P., Ji, G.-P., Sun, G., Cheng, M.-M., Shen, J., and Shao, L.
\newblock Camouflaged object detection.
\newblock In \emph{Proceedings of the IEEE/CVF conference on computer vision and pattern recognition}, pp.\  2777--2787, 2020.

\bibitem[Grill et~al.(2020)Grill, Strub, Altch{\'e}, Tallec, Richemond, Buchatskaya, Doersch, Avila~Pires, Guo, Gheshlaghi~Azar, et~al.]{grill2020bootstrap}
Grill, J.-B., Strub, F., Altch{\'e}, F., Tallec, C., Richemond, P., Buchatskaya, E., Doersch, C., Avila~Pires, B., Guo, Z., Gheshlaghi~Azar, M., et~al.
\newblock Bootstrap your own latent-a new approach to self-supervised learning.
\newblock \emph{Advances in neural information processing systems}, 33:\penalty0 21271--21284, 2020.

\bibitem[Guan et~al.(2024)Guan, Fang, Zhu, Cai, Ling, Yang, and Luo]{guan2024idenet}
Guan, J., Fang, X., Zhu, T., Cai, Z., Ling, Z., Yang, M., and Luo, J.
\newblock Idenet: Making neural network identify camouflaged objects like creatures.
\newblock \emph{IEEE Transactions on Image Processing}, 2024.

\bibitem[He et~al.(2023)He, Dong, Lin, and Lau]{he2023weakly}
He, R., Dong, Q., Lin, J., and Lau, R.~W.
\newblock Weakly-supervised camouflaged object detection with scribble annotations.
\newblock In \emph{Proceedings of the AAAI Conference on Artificial Intelligence}, volume~37, pp.\  781--789, 2023.

\bibitem[He et~al.(2024)He, Xia, Qiao, and Li]{he2024text}
He, Z., Xia, C., Qiao, S., and Li, J.
\newblock Text-prompt camouflaged instance segmentation with graduated camouflage learning.
\newblock In \emph{Proceedings of the 32nd ACM International Conference on Multimedia}, pp.\  5584--5593, 2024.

\bibitem[Hinton(2015)]{hinton2015distilling}
Hinton, G.
\newblock Distilling the knowledge in a neural network.
\newblock \emph{arXiv preprint arXiv:1503.02531}, 2015.

\bibitem[Ho et~al.(2020)Ho, Jain, and Abbeel]{ho2020denoising}
Ho, J., Jain, A., and Abbeel, P.
\newblock Denoising diffusion probabilistic models.
\newblock \emph{Advances in neural information processing systems}, 33:\penalty0 6840--6851, 2020.

\bibitem[Hu et~al.(2024)Hu, Lin, Yan, and Gong]{huleveraging}
Hu, J., Lin, J., Yan, J., and Gong, S.
\newblock Leveraging hallucinations to reduce manual prompt dependency in promptable segmentation.
\newblock In \emph{The Thirty-eighth Annual Conference on Neural Information Processing Systems}, 2024.

\bibitem[Hu et~al.(2023)Hu, Wang, Qin, Dai, Ren, Luo, Tai, and Shao]{hu2023high}
Hu, X., Wang, S., Qin, X., Dai, H., Ren, W., Luo, D., Tai, Y., and Shao, L.
\newblock High-resolution iterative feedback network for camouflaged object detection.
\newblock In \emph{Proceedings of the AAAI Conference on Artificial Intelligence}, volume~37, pp.\  881--889, 2023.

\bibitem[Huang et~al.(2023)Huang, Huang, Yang, Ren, Liu, Li, Ye, Liu, Yin, and Zhao]{huang2023make}
Huang, R., Huang, J., Yang, D., Ren, Y., Liu, L., Li, M., Ye, Z., Liu, J., Yin, X., and Zhao, Z.
\newblock Make-an-audio: Text-to-audio generation with prompt-enhanced diffusion models.
\newblock In \emph{International Conference on Machine Learning}, pp.\  13916--13932. PMLR, 2023.

\bibitem[Jia et~al.(2022)Jia, Yao, Liu, Fan, Liu, and Luo]{jia2022segment}
Jia, Q., Yao, S., Liu, Y., Fan, X., Liu, R., and Luo, Z.
\newblock Segment, magnify and reiterate: Detecting camouflaged objects the hard way.
\newblock In \emph{Proceedings of the IEEE/CVF Conference on Computer Vision and Pattern Recognition}, pp.\  4713--4722, 2022.

\bibitem[Le et~al.(2019)Le, Nguyen, Nie, Tran, and Sugimoto]{le2019anabranch}
Le, T.-N., Nguyen, T.~V., Nie, Z., Tran, M.-T., and Sugimoto, A.
\newblock Anabranch network for camouflaged object segmentation.
\newblock \emph{Computer vision and image understanding}, 184:\penalty0 45--56, 2019.

\bibitem[Lim et~al.(2017)Lim, Son, Kim, Nah, and Mu~Lee]{lim2017enhanced}
Lim, B., Son, S., Kim, H., Nah, S., and Mu~Lee, K.
\newblock Enhanced deep residual networks for single image super-resolution.
\newblock In \emph{Proceedings of the IEEE conference on computer vision and pattern recognition workshops}, pp.\  136--144, 2017.

\bibitem[Loshchilov(2017)]{loshchilov2017decoupled}
Loshchilov, I.
\newblock Decoupled weight decay regularization.
\newblock \emph{arXiv preprint arXiv:1711.05101}, 2017.

\bibitem[Lugmayr et~al.(2022)Lugmayr, Danelljan, Romero, Yu, Timofte, and Van~Gool]{lugmayr2022repaint}
Lugmayr, A., Danelljan, M., Romero, A., Yu, F., Timofte, R., and Van~Gool, L.
\newblock Repaint: Inpainting using denoising diffusion probabilistic models.
\newblock In \emph{Proceedings of the IEEE/CVF conference on computer vision and pattern recognition}, pp.\  11461--11471, 2022.

\bibitem[Lv et~al.(2021)Lv, Zhang, Dai, Li, Liu, Barnes, and Fan]{lv2021simultaneously}
Lv, Y., Zhang, J., Dai, Y., Li, A., Liu, B., Barnes, N., and Fan, D.-P.
\newblock Simultaneously localize, segment and rank the camouflaged objects.
\newblock In \emph{Proceedings of the IEEE/CVF conference on computer vision and pattern recognition}, pp.\  11591--11601, 2021.

\bibitem[Margolin et~al.(2014)Margolin, Zelnik-Manor, and Tal]{margolin2014evaluate}
Margolin, R., Zelnik-Manor, L., and Tal, A.
\newblock How to evaluate foreground maps?
\newblock In \emph{Proceedings of the IEEE conference on computer vision and pattern recognition}, pp.\  248--255, 2014.

\bibitem[Mei et~al.(2021)Mei, Ji, Wei, Yang, Wei, and Fan]{mei2021camouflaged}
Mei, H., Ji, G.-P., Wei, Z., Yang, X., Wei, X., and Fan, D.-P.
\newblock Camouflaged object segmentation with distraction mining.
\newblock In \emph{Proceedings of the IEEE/CVF conference on computer vision and pattern recognition}, pp.\  8772--8781, 2021.

\bibitem[Oquab et~al.(2023)Oquab, Darcet, Moutakanni, Vo, Szafraniec, Khalidov, Fernandez, Haziza, Massa, El-Nouby, et~al.]{oquab2023dinov2}
Oquab, M., Darcet, T., Moutakanni, T., Vo, H., Szafraniec, M., Khalidov, V., Fernandez, P., Haziza, D., Massa, F., El-Nouby, A., et~al.
\newblock Dinov2: Learning robust visual features without supervision.
\newblock \emph{arXiv preprint arXiv:2304.07193}, 2023.

\bibitem[Ouyang \& Key(2021)Ouyang and Key]{ouyang2021maximum}
Ouyang, L. and Key, A.
\newblock Maximum mean discrepancy for generalization in the presence of distribution and missingness shift.
\newblock In \emph{NeurIPS 2021 Workshop on Distribution Shifts: Connecting Methods and Applications}, 2021.

\bibitem[Pang et~al.(2025)Pang, Zhao, Zuo, Zhang, and Lu]{pang2025open}
Pang, Y., Zhao, X., Zuo, J., Zhang, L., and Lu, H.
\newblock Open-vocabulary camouflaged object segmentation.
\newblock In \emph{European Conference on Computer Vision}, pp.\  476--495. Springer, 2025.

\bibitem[Perazzi et~al.(2012)Perazzi, Kr{\"a}henb{\"u}hl, Pritch, and Hornung]{perazzi2012saliency}
Perazzi, F., Kr{\"a}henb{\"u}hl, P., Pritch, Y., and Hornung, A.
\newblock Saliency filters: Contrast based filtering for salient region detection.
\newblock In \emph{2012 IEEE conference on computer vision and pattern recognition}, pp.\  733--740. IEEE, 2012.

\bibitem[Saharia et~al.(2022)Saharia, Ho, Chan, Salimans, Fleet, and Norouzi]{saharia2022image}
Saharia, C., Ho, J., Chan, W., Salimans, T., Fleet, D.~J., and Norouzi, M.
\newblock Image super-resolution via iterative refinement.
\newblock \emph{IEEE transactions on pattern analysis and machine intelligence}, 45\penalty0 (4):\penalty0 4713--4726, 2022.

\bibitem[Song et~al.(2023)Song, Kang, Wei, Liu, Dian, and Li]{song2023fsnet}
Song, Z., Kang, X., Wei, X., Liu, H., Dian, R., and Li, S.
\newblock Fsnet: Focus scanning network for camouflaged object detection.
\newblock \emph{IEEE Transactions on Image Processing}, 32:\penalty0 2267--2278, 2023.

\bibitem[Sun et~al.(2022)Sun, Wang, Chen, and Xiang]{sun2022boundary}
Sun, Y., Wang, S., Chen, C., and Xiang, T.~Z.
\newblock Boundary-guided camouflaged object detection.
\newblock In \emph{31st International Joint Conference on Artificial Intelligence, IJCAI 2022}, pp.\  1335--1341. International Joint Conferences on Artificial Intelligence, 2022.

\bibitem[Wang et~al.(2020)Wang, Li, Zhu, Tian, and Shan]{wang2020dual}
Wang, L., Li, D., Zhu, Y., Tian, L., and Shan, Y.
\newblock Dual super-resolution learning for semantic segmentation.
\newblock In \emph{Proceedings of the IEEE/CVF conference on computer vision and pattern recognition}, pp.\  3774--3783, 2020.

\bibitem[Wang et~al.(2024)Wang, Yang, Zhang, Wang, and Zheng]{wang2024depth}
Wang, L., Yang, J., Zhang, Y., Wang, F., and Zheng, F.
\newblock Depth-aware concealed crop detection in dense agricultural scenes.
\newblock In \emph{Proceedings of the IEEE/CVF Conference on Computer Vision and Pattern Recognition}, pp.\  17201--17211, 2024.

\bibitem[Wang et~al.(2023)Wang, Yang, Yu, Wang, Chen, and Zheng]{wang2023depth}
Wang, Q., Yang, J., Yu, X., Wang, F., Chen, P., and Zheng, F.
\newblock Depth-aided camouflaged object detection.
\newblock In \emph{Proceedings of the 31st ACM International Conference on Multimedia}, pp.\  3297--3306, 2023.

\bibitem[Wei et~al.(2020)Wei, Wang, and Huang]{wei2020f3net}
Wei, J., Wang, S., and Huang, Q.
\newblock F$^3$net: fusion, feedback and focus for salient object detection.
\newblock In \emph{Proceedings of the AAAI conference on artificial intelligence}, volume~34, pp.\  12321--12328, 2020.

\bibitem[Yue et~al.(2024)Yue, Wang, and Loy]{yue2024resshift}
Yue, Z., Wang, J., and Loy, C.~C.
\newblock Resshift: Efficient diffusion model for image super-resolution by residual shifting.
\newblock \emph{Advances in Neural Information Processing Systems}, 36, 2024.

\bibitem[Zhang et~al.(2023)Zhang, Rao, and Agrawala]{zhang2023adding}
Zhang, L., Rao, A., and Agrawala, M.
\newblock Adding conditional control to text-to-image diffusion models.
\newblock In \emph{Proceedings of the IEEE/CVF International Conference on Computer Vision}, pp.\  3836--3847, 2023.

\bibitem[Zhu et~al.(2022)Zhu, Li, Xie, Yan, Liang, Chen, Wei, and Qin]{zhu2022can}
Zhu, H., Li, P., Xie, H., Yan, X., Liang, D., Chen, D., Wei, M., and Qin, J.
\newblock I can find you! boundary-guided separated attention network for camouflaged object detection.
\newblock In \emph{Proceedings of the AAAI conference on artificial intelligence}, volume~36, pp.\  3608--3616, 2022.

\end{thebibliography}
\bibliographystyle{icml2025}

\end{document}